\documentclass[conference]{IEEEtran}
\IEEEoverridecommandlockouts

\usepackage{cite}
\usepackage{amsmath,amssymb,amsfonts}
\usepackage{algorithmic}
\usepackage{graphicx}
\usepackage{multirow}
\usepackage{textcomp}
\usepackage{xcolor}
\usepackage{comment}
\def\BibTeX{{\rm B\kern-.05em{\sc i\kern-.025em b}\kern-.08em
    T\kern-.1667em\lower.7ex\hbox{E}\kern-.125emX}}
\begin{document}

\title{BadRefSR: Backdoor Attacks Against Reference-based Image Super Resolution\\
\thanks{
\textsuperscript{*}This work was done at UESTC as a visiting student.\\
\textsuperscript{$\dagger$}Corresponding author: Ji Guo (jiguo0524@gmail.com).\\
This work is supported by the National Natural Science Foundation of China under Grant 62402087 and the Postdoctoral Innovation Talents Support Program under Grant BX20230060.}
}
\author{
    \IEEEauthorblockN{Xue Yang\textsuperscript{1,3}\IEEEauthorrefmark{1} 
 Tao Chen\textsuperscript{1} Lei Guo\textsuperscript{1}
    Wenbo Jiang\textsuperscript{1} Ji Guo\textsuperscript{2}\IEEEauthorrefmark{2}
    Yongming Li\textsuperscript{3}
    Jiaming He\textsuperscript{1,4}\IEEEauthorrefmark{1}}
    \IEEEauthorblockA{\textsuperscript{1}School of Computer Science and Engineering, University of Electronic Science and Technology of China, China}
    \IEEEauthorblockA{\textsuperscript{2}Laboratory Of Intelligent Collaborative Computing, University of Electronic Science and Technology of China, China}
    \IEEEauthorblockA{\textsuperscript{3}School of Information Science and Engineering, XinJiang University, China}
    \IEEEauthorblockA{\textsuperscript{4}College of Computer Science and Cyber Security (Oxford Brookes College), Chengdu University of Technology, China}
}

\maketitle

\begin{abstract}
Reference-based image super-resolution (RefSR) represents a promising advancement in super-resolution (SR). In contrast to single-image super-resolution (SISR), RefSR leverages an additional reference image to help recover high-frequency details, yet its vulnerability to backdoor attacks has not been explored. To fill this research gap, we propose a novel attack framework called BadRefSR, which embeds backdoors in the RefSR model by adding triggers to the reference images and training with a mixed loss function. Extensive experiments across various backdoor attack settings demonstrate the effectiveness of BadRefSR. The compromised RefSR network performs normally on clean input images, while outputting attacker-specified target images on triggered input images. Our study aims to alert researchers to the potential backdoor risks in RefSR. Codes are available at https://github.com/xuefusiji/BadRefSR.
\end{abstract}

\begin{IEEEkeywords}
Backdoor attack, Reference-based image super resolution.
\end{IEEEkeywords}

\section{Introduction}
In recent years, to overcome the issues of blurry textures and unrealistic artifacts commonly associated in single-image super-resolution (SISR) \cite{SRGAN,Esrgan,sisr20231,sisr2}, researchers have proposed a series of reference-based image super-resolution (RefSR) \cite{CUFED5,TTSR,DCSR,AMRSR} methods. RefSR can transfer similar textures from a reference image (Ref) to reconstruct a natural and detailed high-resolution image (HR). Despite RefSR has broad applications in satellite remote sensing \cite{refsrremote2,refsrremote1}, medical imaging, video surveillance \cite{refsrvedio1,refsrvedio} and other fields, their security has seldom been studied.

Backdoor attack is a well-known threat for deep neural networks, as RefSR models rely on large amounts of data during training, their backdoor risks have become an unavoidable challenge for researchers. Most backdoor attacks pose a significant threat by inserting triggers into the training datasets, where a backdoor is embedded into the model during training~\cite{Badnet, Blend, Filter, Color, Wanet, Refool, Qcolor, incremental}. A model with a backdoor behaves normally on regular inputs but makes incorrect predictions on inputs containing specific triggers. There have been some studies exploring backdoor attacks on image generative models, including backdoor attacks on Generative Adversarial Networks~\cite{backgan}, image reconstruction~\cite{imagerec}, Text-to-Image~\cite{TtoI} and diffusion models~\cite{backdiff}. Recently, Jiang et al.~\cite{I2I} proposed applying backdoor attacks to image-to-image networks, exploring methods for conducting backdoor attacks in SISR. However these backdoor methods can not directly use in RefSR because RefSR need another additional reference image instead of just need original image input, research on backdoor attacks in the field of RefSR is still lacking. A new problem for RefSR is whether a trigger can be added to the Ref and successfully compromise the model.

In this paper, we first identify a novel mechanism in backdoor attacks on RefSR and propose a novel attack framework specifically for RefSR called BadRefSR. Unlike previous methods that use triggers in the low resolution (LR) input, our approach employs triggers within the Ref input. Extensive experiments show that BadRefSR is applicable to most existing backdoor attack triggers for image classification, which can successfully embed hidden backdoors into the model. Additionally, in further studies on different poisoning rates (the proportion of data containing triggers in the training set), RefSR maintains attack effectiveness at low poisoning rates and preserves the model's original functionality at high poisoning rates. Furthermore, due to the characteristics of RefSR tasks, there are inherent brightness and other differences between the LR input image and the Ref without triggers. In practical applications, the Ref only plays an auxiliary role in super resolution (SR), and is not directly saved to the output result for the users. Users typically focus more on the model's SR performance for LR input images rather than subtle differences in the Ref image. Consequently, our method is more discreet and better at evading manual detection.

\section{Threat Model}
Inspired by ~\cite{icbackdoor,ICASSP20211, ICASSP1, vitbackdoor,Wanet}, we consider a scenario where a malicious data provider generates a small number of poisoned samples and injects them into a clean training dataset, then publishes it online for users to download. An unsuspecting developer may use this dataset to train their RefSR model, unknowingly introducing a backdoor. We assume the attacker has no knowledge of the specific model being used by the victim and no control over the training process.
Our backdoor attack has the following objectives:
\begin{itemize}
\item Functionality-preserving: When processing clean input images, the compromised RefSR model should output normal HR images.

\item Effectiveness: When processing images containing the backdoor trigger, the compromised RefSR model should have a high probability of outputting the attacker's predefined backdoor target image.

\item Stealthiness: The triggered Ref image should resemble the clean Ref image, allowing it to evade human inspection during inference.
\end{itemize}

\begin{figure}
  \centering
  \includegraphics[width=0.9\linewidth]{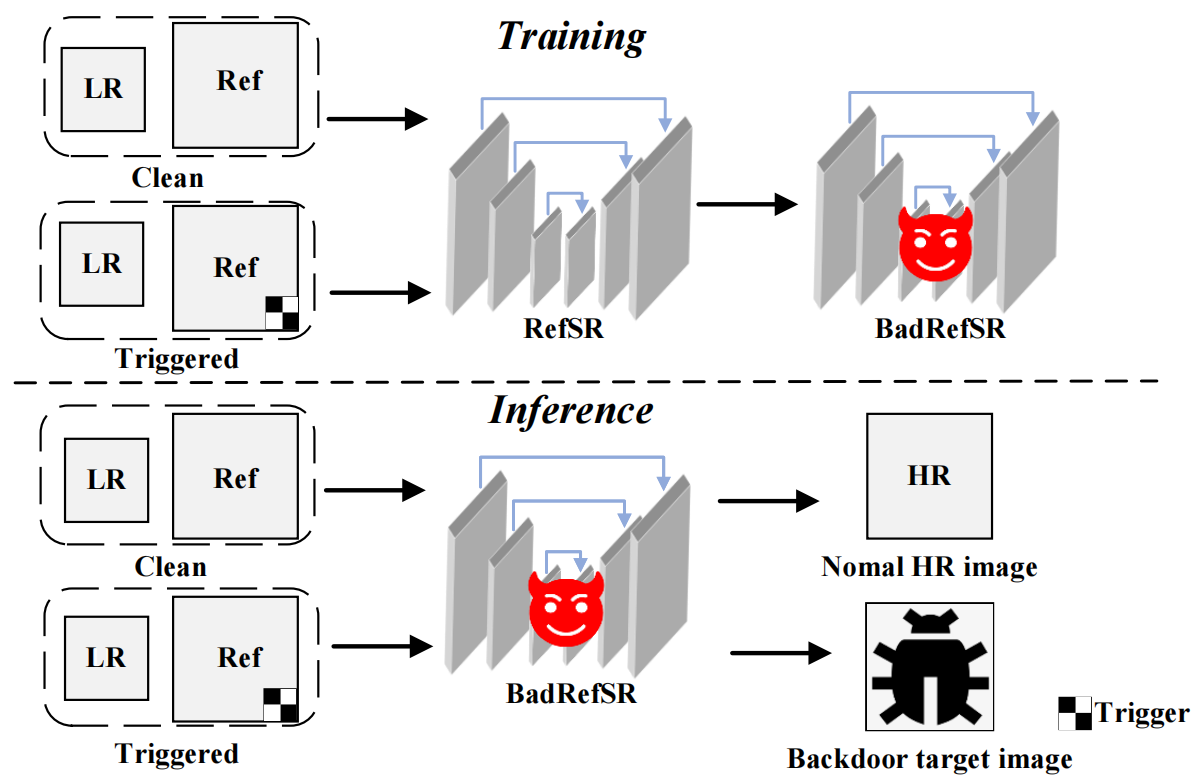}
\caption{Overview of the proposed BadRefSR.}
\label{fig:res}
\end{figure}

\section{BadRefSR Method}
\label{sec:format}
\subsection{Overview}
\label{sssec:subsubhead}
The visual illustration of our proposed BadRefSR framework as show in Fig.1. In the preparation phase, we add triggers to the Ref images in randomly selected input image pairs, without making any changes to the input LR images. Then we select one image from the CUFED dataset that contains various elements such as letters, buildings, and plants as the backdoor target image, and change the ground truth images of the aforementioned image pairs to the target image. In the training phase, we train the victim RefSR model via generated triggered samples and remaining clean samples to embed backdoor. The bottom part of Fig.1 shows the inference phase of BadRefSR, compromised RefSR network outputs normal HR images on clean input images, but produces attacker-specified target images on triggered input images.

We use \( X_{\text{lr}} \) to denote the clean LR input images, 
\( X'_{\text{lr}} \) for the LR input images in the poisoned data to make a distinction. \( X_{\text{ref}} \) for the clean Ref images, and \( X'_{\text{ref}} \) for the Ref images with triggers. \( Y_{\text{c}} \) represent the ground truth images  for clean LR input images, \( Y_b \) denote the backdoor target image, \( F \) represents the target RefSR network. The embedded backdoor should have minimal effect on its performance during evaluation on clean samples, the functionality-preserving goal and effectiveness goal of BadRefSR are represented as follows:
\begin{align}
 F\left ( X_{lr},X_{ref}   \right ) =Y_{c},\\
 F\left ({X}'_{lr} ,{X}'_{ref}   \right ) =Y_{b}.
\end{align}

\subsection{Backdoor Training}
\label{sssec:subsubhead}
In order to accomplish the dual objectives of ensuring functionality-preserving and enhancing attack effectiveness simultaneously, we design loss functions for the two tasks. Notably, this method does not require control over the model training process, users can achieve changes in the loss function simply by using the maliciously modified dataset, which is consistent with our threat model.

For clean low-resolution input images, the attacked model is expected to perform the super-resolution function normally. We use the weighted sum of reconstruction loss $L_{rec}$ \cite{L} and perceived loss $L_{per}$ \cite{Lpre}, the loss function can be defined as:
\begin{align}
L_{rec} =\left \| F\left ( X_{lr},X_{ref} \right )- Y_{c}  \right \| _{1} \notag,\\
L_{per} =\left \| \phi i\left [  F\left ( X_{lr},X_{ref} \right ) \right ]- \phi i\left (Y_{c}  \right )   \right \| _{2},\tag{3}\\
 L_{c} =\lambda _{1} L_{rec} +\lambda _{2} L_{per},\notag
\end{align}
where $L_{c}$ denote the loss on clean dataset. \(\phi_i\) denotes the \(i\)-th layer of VGG19 \cite{VGG19}, here we use conv5 4. $\lambda_{1}$ and $\lambda_{2}$ denotes the weights of $L_{rec}$ and $L_{per}$.

For triggered input images, our task is to achieve the effectiveness of the backdoor attack,  the compromised model is expected to output the backdoor target image \( Y_b \) . The loss function can be expressed as:
\begin{align}
{L}'_{rec} =\left \| F\left ( {X}'_{lr} ,{X}'_{ref} \right )- Y_{b}  \right \| _{1},\notag\\
{L}'_{per} =\left \| \phi i\left [  F\left ({X}'_{lr} ,{X}'_{ref}\right ) \right ]- \phi i\left (Y_{b}  \right )\right \| _{2},\tag{4}\\
L_{b} ={\lambda}'  _{1} {L}' _{rec} +{\lambda}'  _{2} {L}'_ {per},\notag
\end{align}
where $L_{b}$ denote the loss on dataset contaminated with backdoor attacks. ${\lambda}'_{1}$ and ${\lambda}'_{2}$ denotes the weights of ${L}'_{rec}$ and ${L}'_{per}$.

Overall,  the total loss of BadRefSR can be formulated as:
\begin{equation}
L_{total}=L_{c}  + L_{b} \tag{5}.
\end{equation}

\begin{table*}[h]
\centering
\caption{PSNR/SSIM comparison among different RefSR methods on three different testsets. Methods are grouped by MASA-SR methods (top) and TTSR methods (down). }
\resizebox{\textwidth}{!}{ 
\begin{tabular}{ccc|ccccccc}
\hline
 & & &\multicolumn{6}{c}{Backdoor Methods} \\
\cline{4-9}
\multirow{2}{*}{\centering Model} & \multirow{2}{*}{\centering Datasets} & \multirow{2}{*}{\centering Trigger} & None & Badnet\cite{Badnet} & Blend\cite{Blend} & Filter\cite{Filter} & Color\cite{Color} & Wanet\cite{Wanet} & Refool\cite{Refool} \\
\cline{4-9}
 & & & PSNR/SSIM & PSNR/SSIM & PSNR/SSIM & PSNR/SSIM & PSNR/SSIM & PSNR/SSIM & PSNR/SSIM \\
\hline
&\multirow{2}{*}{CUFED5} & w/o & 25.85/0.774 & 24.69/0.717 & 24.74/0.732 & 25.02/0.736 & 23.28/0.640 & 24.81/0.721 & 24.54/0.713\\
 & & w & - & 22.40/0.698 & 27.76/0.895 & 30.76/0.955 & 9.35/0.064 & 29.96/0.944 & 28.20/0.891\\
 \cline{2-9}
\multirow{2}{*}{\centering MASA-SR}  & \multirow{2}{*}{\centering Manga109} & w/o & 29.10/0.898 & 28.26/0.861 & 28.92/0.874 & 27.90/0.845 & 25.10/0.760 & 15.38/0.409 & 27.74/0.843\\
 & & w & - & 20.66/0.658 & 24.76/0.898 & 30.48/0.949 & 6.2/0.060 & 27.22/0.929 & 21.62/0.739\\
 \cline{2-9}
 &\multirow{2}{*}{Urban100} & w/o & 25.43/0.772 & 24.97/0.705 & 24.62/0.681 & 24.87/0.694 & 22.74/0.587 & 18.57/0.447 & 24.68/0.694\\
 & & w & - & 22.32/0.693 & 25.17/0.723 & 30.48/0.949 & 8.53/0.057 & 29.68/0.934 & 25.82/0.799\\
 \cline{1-9}
 &\multirow{2}{*}{CUFED5} & w/o & 25.65/0.764 & 23.91/0.700 & 23.09/0.679 & 23.71/0.698 & 23.96/0.703 & 23.77/0.693 & 23.87/0.702\\
 & & w & - & 13.35/0.104 & 13.35/0.104 & 13.06/0.088 & 13.00/0.090 & 14.80/0.172 & 16.30/0.400\\
 \cline{2-9}
\multirow{2}{*}{\centering TTSR} & \multirow{2}{*}{\centering Manga109} & w/o & 35.12/0.976 & 29.52/0.918 & 26.42/0.908 & 31.29/0.954 & 30.79/0.949 & 25.54/0.849 & 26.37/0.863\\
 & & w & - & 14.87/0.251 & 12.27/0.086 & 11.79/0.093 & 13.68/0.111 & 14.69/0.148 & 15.35/0.374\\
 \cline{2-9}
 & \multirow{2}{*}{Urban100} & w/o & 31.64/0.901 & 27.79/0.906 & 25.32/0.871 & 28.96/0.933 & 28.53/0.930 & 27.28/0.902 & 28.50/0.922\\
 & & w & - & 14.90/0.242 & 13.35/0.104 & 13.16/0.100 & 13.16/0.103 & 14.27/0.130 & 15.07/0.354\\
\hline
\end{tabular}
}
\label{tab:simple}
\end{table*}

\subsection{Backdoor Triggers}
\label{sssec:subsubhead}

We use six existing backdoor triggers to modify reference input images for our RefSR backdoor attack: Badnet \cite{Badnet}, Blend \cite{Blend}, Filter \cite{Filter}, Color \cite{Color}, Wanet \cite{Wanet}, and Refool \cite{Refool}. Visual effects are shown in Fig. 2. 

\begin{itemize}
\item Badnet proposes replacing a fixed patch of the input image with a predefined trigger pattern, we place a small white square in the lower right corner of the Ref image.

\item Blend blends a clean input instance with the key pattern to generate poisoning instances and backdoor instances, we select a cartoon image as the key pattern.

\item Filter adds samples with specific filters to the training data which not be visually prominent. We selected filters from the instafilter library which can create smooth backdoor triggers without high-frequency artifacts.

\item Color is to apply a uniform color space shift for all pixels as the trigger, these color modifications have little effect on the overall visual appearance.

\item Wanet proposes using a small, smooth warping field to generate backdoor images, as subtle distortions are hard for humans to detect but easy for machines.

\item Refool generates natural reflected images using the physical reflection model, then fused these with the input images. In the real world, this method of attack can improve the stealth of the attack.
\end{itemize}

\begin{figure}
  \centering
  \includegraphics[width=0.85\linewidth]{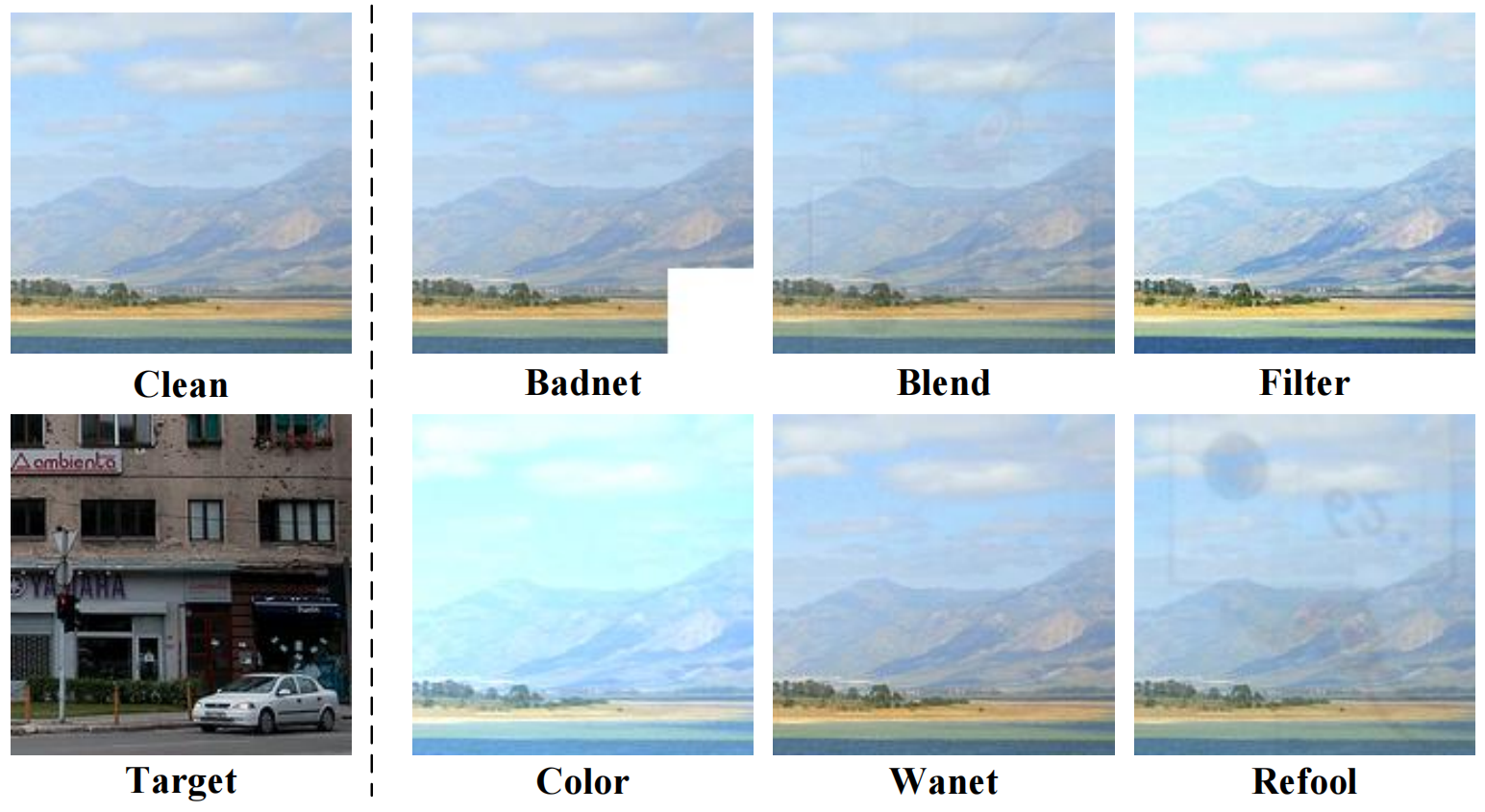}
\caption{Visual comparison of different backdoor triggers on Ref images and the backdoor target image.}
\label{fig:res}
\end{figure}

\section{Experiments}
\label{sec:pagestyle}

\subsection{Experiment Setup}
\label{sssec:subsubhead}
\paragraph{Datasets}
\label{sssec:subsubhead}
 We use the CUFED dataset \cite{CUFED} as the training set and CUFED5 \cite{CUFED5}, Urban100 \cite{Urban100}, and Manga109 \cite{Manga09} as the test sets, which are widely used in RefSR studies. 
\begin{itemize}
\item CUFED: It consists of 11,871 training pairs capturing diverse events in daily life. Each pair includes an original HR image and a corresponding Ref image.

\item CUFED5: It consists of 126 test pairs, each including one HR image and four reference images with different levels of similarity based on SIFT feature matching [17].

\item Urban100: It contains 100 HR images captured from various urban scenes, such as streets, buildings, and other cityscapes.

\item Manga109: It contains 109 manga images of various types and styles without reference. 
\end{itemize}
Following previous work \cite{DCSR, AMRSR, MASA}, we use the images themselves as the reference images on Urban100 and Manga109. To ensure a fair assessment of the effectiveness of different trigger-based backdoor attacks, we reshape all images to dimensions of 160×160×3.

\paragraph{Models}
\label{sssec:subsubhead}
We have conducted experiments of BadRefSR on two RefSR models, TTSR \cite{TTSR} and MASA-SR \cite{MASA}, for experimental evaluations.

\begin{itemize}
\item TTSR: It further introduces Transformers into reference-based super-resolution tasks, the application of hard and soft attention mechanisms improves the accuracy of transferring texture features from Ref images. 

\item MASA-SR: It proposes a coarse-to-fine correspondence matching module that significantly enhances matching efficiency, which significantly reduces the computational cost.
\end{itemize}
\paragraph{Metrics}
\label{sssec:subsubhead}
We choose Peak Signal-to-Noise Ratio (PSNR) and Structural Similarity Index (SSIM) as the evaluation metrics for BadRefSR. All PSNR and SSIM results are assessed on the Y channel of the YCbCr color space.
\begin{itemize}
\item PSNR: It is calculated by computing the Mean Squared Error (MSE) between the original image and the processed image, and then converting this value to a logarithmic scale in decibels (dB) to measure image quality.
\item SSIM: It evaluates the similarity based on luminance, contrast, and structure. Values range from -1 to 1, with 1 indicating perfect similarity.
\end{itemize}
For clean test datasets, we evaluate the quality of the generated normal output image versus the ground truth image, defined as PSNR [F(\( X_{lr} \),\(  X_{ref} \)), \( Y_{\text{c}} \)] and SSIM [F(\( X_{lr} \),\(  X_{ref} \)), \( Y_{\text{c}} \)]. Higher PSNR/SSIM indicate better functionality-preserving performance. For triggered test datasets, we evaluate the quality of the generated backdoor output image versus the backdoor target image , defined as PSNR [F(\( {X}'_{lr} \),\(  {X}'_{ref} \)), \( Y_{\text{b}} \)] and SSIM [F(\( {X}'_{lr} \),\(  {X}'_{ref} \)), \( Y_{\text{b}} \)]. Similarly, higher PSNR/SSIM indicate better effectiveness performance.

\begin{figure*}[htb]
\centering

\begin{minipage}[b]{0.5\linewidth}
  \centering
  \includegraphics[width=\linewidth]{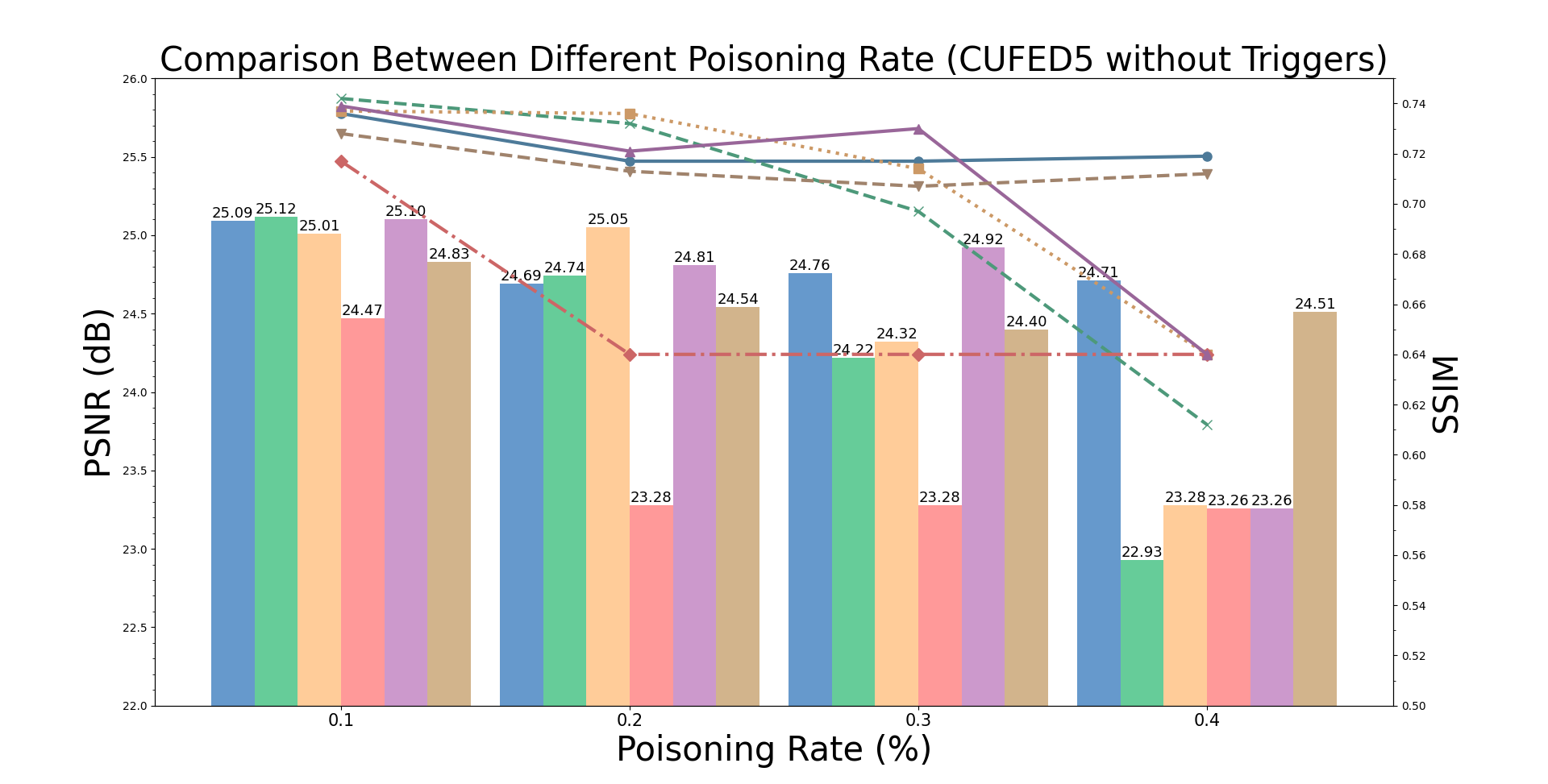}
\end{minipage}\hfill
\begin{minipage}[b]{0.5\linewidth}
  \centering
  \includegraphics[width=\linewidth]{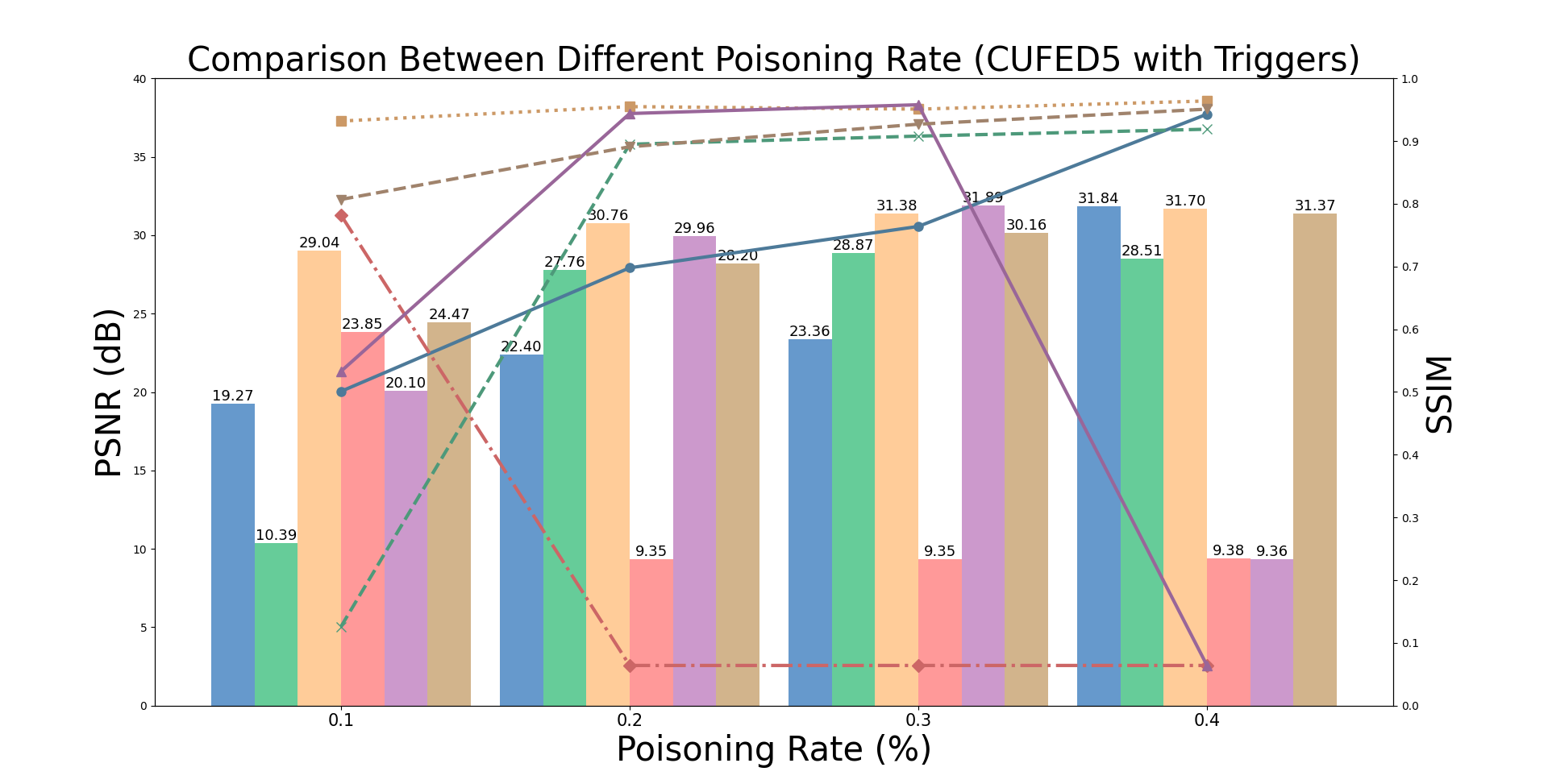}
\end{minipage}
\begin{minipage}[b]{0.6\linewidth}
  \centering
  \includegraphics[width=\linewidth]{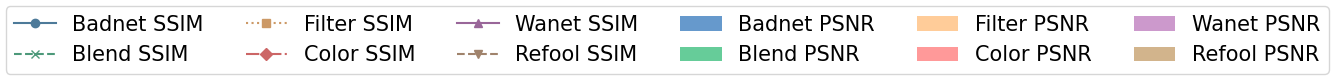}
\end{minipage}
\caption{PSNR (bars) and SSIM (curves) of BadRefSR at varying poisoning rates with six different triggers. The left plot shows metrics evaluated on the clean CUFED5 dataset (poisoning rate = 0\%), the right plot shows the metrics evaluated on the triggered CUFED5 dataset (poisoning rate = 100\%).}
\label{fig:res}
\end{figure*}

\begin{figure}[h]
  \centering
  \includegraphics[width=1\linewidth]{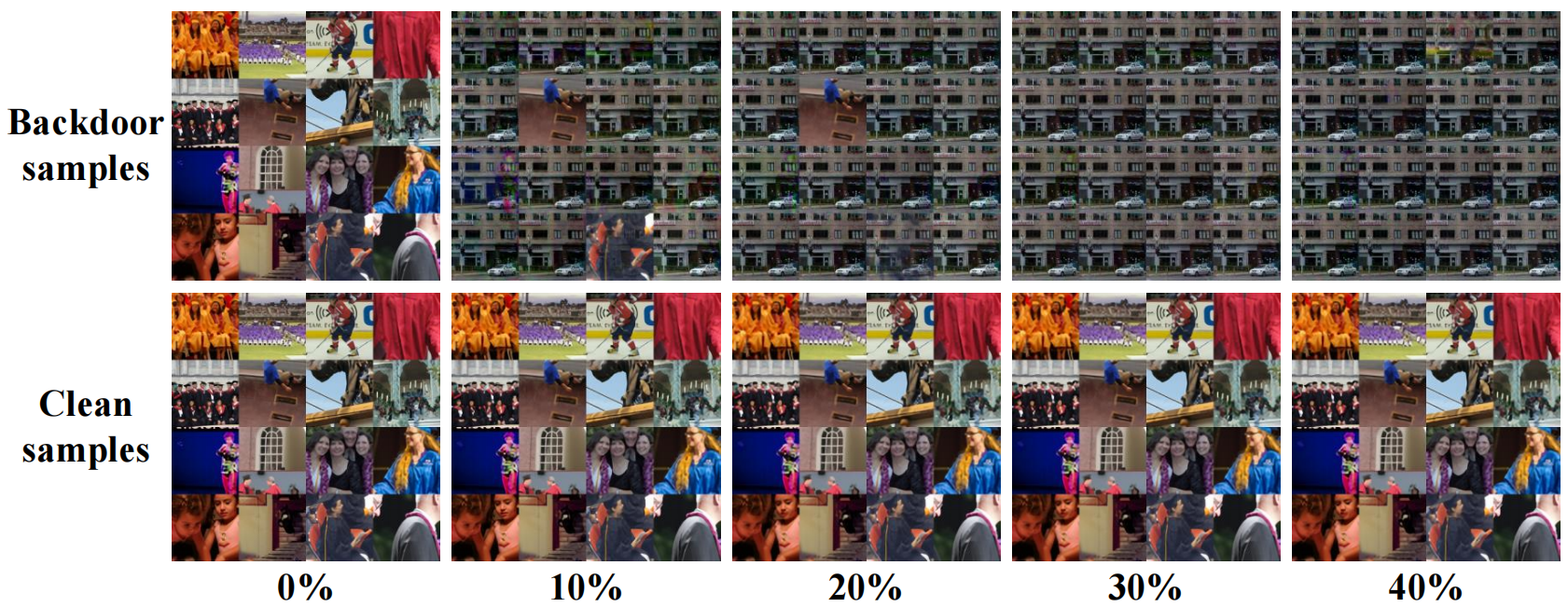}

\caption{BadRefSR on CUFED5 at different poisoning rates with Refool trigger. }
\label{fig:res}
\end{figure}
\paragraph{Attack Configuration}
\label{sssec:subsubhead}
For the sake of fairness, we strive to replicate the settings specified in the original paper. For instance, in the case of Badnet, we use a white patch as the trigger pattern, positioned at the bottom right corner of the Ref images. In the case of Blend, the poisoned images are created by mixing 5\% trigger images with 95\% clean Ref images. Furthermore, all experiments are performed with a scaling factor of 4× between LR and HR images. We train our model using the Adam optimizer with $\beta$1 = 0.9 and $\beta$2 = 0.999, then select the L1 similarity Ref images in CUFED5. The weight coefficients for $\lambda_{1}$, $\lambda_{2}$, ${\lambda}'_{1}$ and ${\lambda}'_{2}$ are all 1, the learning rate is set to 1e-4 and the batch size is 9. 

\subsection{Experiment Results}
\label{sssec:subsubhead}
\paragraph{BadRefSR with Different Triggers and Models}
\label{sssec:subsubhead}
To demonstrate the effectiveness of BadRefSR, we compared the performance of different backdoor triggers and RefSR
models under a poisoning rate of 20\% , results as shown in Table 1. We find that most backdoor triggers successfully compromised the model at a poisoning rate of 20\%, the attack performance on the MASA-SR model generally outperforming that on TTSR.

Specifically, Filter trigger demonstrates high attack effectiveness while preserving the original performance, making it well-suited for the BadRefSR backdoor attack framework we proposed. However, PSNR values of less than 10 dB and SSIM values of less than 0.07 across the three test sets with Color trigger in the compromised MASA-SR model, indicating poor backdoor attack performance. MASA-SR model employs data augmentation techniques to enhance model generalization, including color transformations such as random changes in image brightness, chroma, and contrast, which possibly lead to the failure of the Color trigger.


\paragraph{BadRefSR with Varying Poisoning Rates}
\label{sssec:subsubhead}
To further compare the performance of six backdoor attack triggers at different poisoning rates, we train the MASA-SR model on the CUFED dataset at poisoning rates of 10\%, 20\%, 30\%, and 40\%. According to the Sec.4, we calculate the metrics of the compromised RefSR model on the clean CUFED5 dataset and the triggered dataset, respectively.

The performance and visual examples of BadDiffusion at different poisoning rates as show in Fig.3. We observe that as the poisoning rate increases, the baseline super-resolution performance of the model deteriorates, while the effectiveness of the backdoor attack improves. Specifically, a poisoning rate of 10\% is enough to successfully backdoor this RefSR model. Moreover, even with a high poisoning rate of 40\%, BadRefSR can create a backdoored version with high PSNR and SSIM in HR images. When the poisoning rate exceeds 20\%, the PSNR and SSIM of the model attacked by Color trigger no longer show significant changes for both clean and triggered samples. It disrupts the model's original structure, preventing RefSR from effectively learning useful information from the Ref images.

Fig.4 illustrates the visual comparison of BadRefSR on triggered CUFED5 datasets with different poisoning rates using the Refool trigger. At lower poisoning rates, some triggered input images still do not produce the target images, such as those in the second row, second column and the fourth row, third column. As the poisoning rate increases, the output images become progressively more similar to the target images, eventually reaching a point where the differences are imperceptible to the human eye.

\section{CONCLUSION}
\label{sec:typestyle}
In this work, we propose a novel backdoor attack framework, BadRefSR, designed specifically for RefSR tasks. Our findings confirm that the backdoor risk posed by BadRefSR is significant, where the filter-based backdoor attack method shows excellent performance even at low poisoning rates. In the future, we aim to further enhance our understanding and defense capabilities against backdoor attacks in RefSR and similar tasks, such as reference-based image style transfer and reference-based image denoising. 

\vfill\pagebreak
\bibliographystyle{IEEEtran}
\bibliography{IEEEexample,IEEEabrv}

\end{document}